# Deep Convolutional Neural Networks for Map-Type Classification

Xiran Zhou, Wenwen Li, Samantha T. Arundel, Jun Liu

**ABSTRACT:** Maps are an important medium that enable people to comprehensively understand the configuration of cultural activities and natural elements over different times and places. Although massive maps are available in the digital era, how to effectively and accurately access the required map remains a challenge today. Previous works partially related to map-type classification mainly focused on map comparison and map matching at the local scale. The features derived from local map areas might be insufficient to characterize map content. To facilitate establishing an automatic approach for accessing the needed map, this paper reports our investigation into using deep learning techniques to recognize seven types of map, including topographic map, terrain map, physical map, urban scene map, the National Map, 3D map, nighttime map, orthophoto map, and land cover classification map. Experimental results show that the state-of-the-art deep convolutional neural networks can support automatic map-type classification. Additionally, the classification accuracy varies according to different map-types. We hope our work can contribute to the implementation of deep learning techniques in cartographical community and advance the progress of Geographical Artificial Intelligence (GeoAI).

# Introduction

Maps are an important medium that enable people to comprehensively understand the configuration of cultural activities and natural elements over different times and places. Map features, such as text and geographical features, are used to benefit the representation and communication in cartography and the GIScience community (Lloyd and Bunch, 2003). In the last two decades, Internet and spin-off techniques have significantly changed the nature of map generation and the use of maps (Hurst and Clough, 2013). Due to the advent of web-based service technologies, cyberinfrastructure, and volunteered geographic information (Li, Yang and Yang, 2010), a number of online platforms and tools such as Google Maps, Bing Maps, and Wikimapia are available for map creation, visualization, and geospatial analysis. Currently, maps are not used by geographical domain experts to conduct geospatial computing and analysis. The information included in maps allows the public to better facilitate daily activities such as ridesharing, delivery, and transportation network analysis, to name just a few.

Although a great number of maps are available in the digital era, how to effectively and accurately access the required map remains a challenge today. Three problems have left this challenge unsolved. First, a majority of maps available from the Internet lack map elements like map title, direction indicators, or legends, leaving the reader to rely only on the map frame itself in order to interpret the content. Second, new techniques enable the creation of immense map repositories containing maps with diverse themes, configurations and designs. This diversity increases the difficulty in accessing appropriate maps. Third, unlike the objects in a photograph or image, which are easily characterized, it is impossible to precisely characterize maps that are short their defining map elements. For example, topographic maps and road maps may both contain road networks and streets.

To our knowledge, no literature with respect to map-type classification has yet been reported. Previous works partially related to map-type classification mainly focused on map comparison and map matching at the local scale (Power, Simms and White, 2001; Li and Huang, 2002; Fritz and See, 2005; Zhu, Y., et al., 2017). The local map scales defined by these approaches include pixels, pixel blocks (or superpixels), and polygons (or map objects) (Stehman and Wickham, 2011). However, features derived from local map areas might be insufficient to characterize map content, since overlaps can always be observed in different types of maps. For instance, it is possible to observe water in both ocean maps and topographic maps. An automated approach ensures the availability of needed maps and is essential to facilitate the role of maps in geographical analysis and public activities. Thus, this paper reports our investigation into using deep learning techniques to recognize different types of map.

# Method

## *Dataset*

The datasets for map-type classification are selected from a benchmark called *deepMap*. We created this benchmark to provide datasets for studying automatic map classification with deep learning techniques. All data in *deepMap* are collected from the online maps of ArcGIS, Google Maps, the National Map of the USGS, and other online search engines. *deepMap* offers three types of benchmark datasets: a map text dataset, a text-labeled map dataset, and a map dataset. In this paper, we use the map dataset to conduct map-type classification with Deep Convolutional Neural Networks (DCNNs). The dimensionality of each image in the map dataset is 256×256×3, where 256×256 denotes image size and 3 refers to the RGB channels of an image. *deepMap* contains seven available map categories: 1) topographic map/terrain map/physical map, 2) urban scene map, 3) the National Map, 4) 3D map, 5) nighttime map, 6) orthophoto map, and 7) land cover classification map (Figure 1). Each map category contains around 200 maps in total, and the total number of maps in the map dataset is 1812.

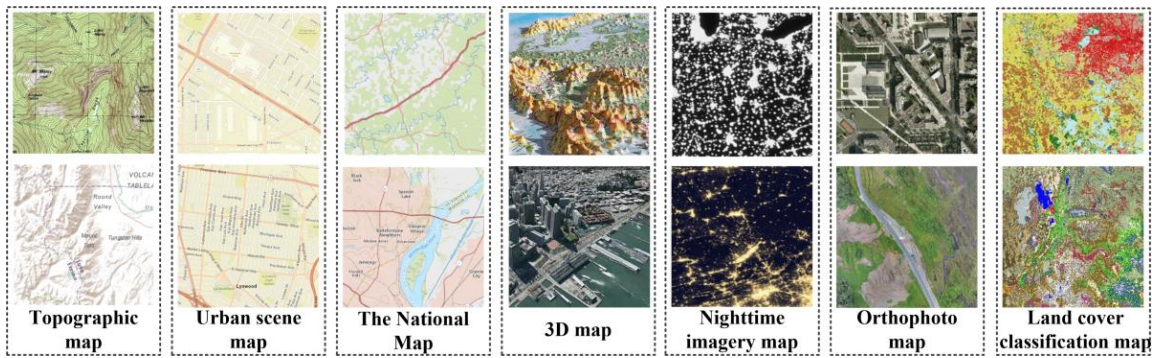

Figure 1: Illustration of the seven types of maps found in the *deepMap* dataset.

## Deep convolutional neural networks

Previous machine learning (ML) approaches, or "shallow ML," have foundered when handling complex functions and features, and generally require substantial labor in training data to obtain satisfactory results (LeCun, Bengio and Hinton, 2015). Deep learning (DL) approaches enable computers to spontaneously access highly valuable information through unsupervised learning, and discover the high-level representations of data based on a multi-layered processing framework. In the last five years, a large number of DCNNs have produced impressive image classifications. Thus, we intend to apply DCNNs to classify map-types based on the map content when metadata and other auxiliary information (map title, map legend, etc.) are not available. Figure 2 shows a general architecture of a DCNN.

A DCNN is a class of multi-layered neural networks designed to exploit features of image or speech signals, which generally consist of two parts: feature generation and classification. The input image is an image that has RGB channels. Feature generation includes a number of convolutional layers and pooling layers (or unsampling layers) to produce a feature map that includes high-level representation of the input image. Using the resulting convolutional layer, the classification includes fully-connected or densely-connected layer(s) and a classifier (e.g. softmax) to classify the input image as one of the predefined classes.

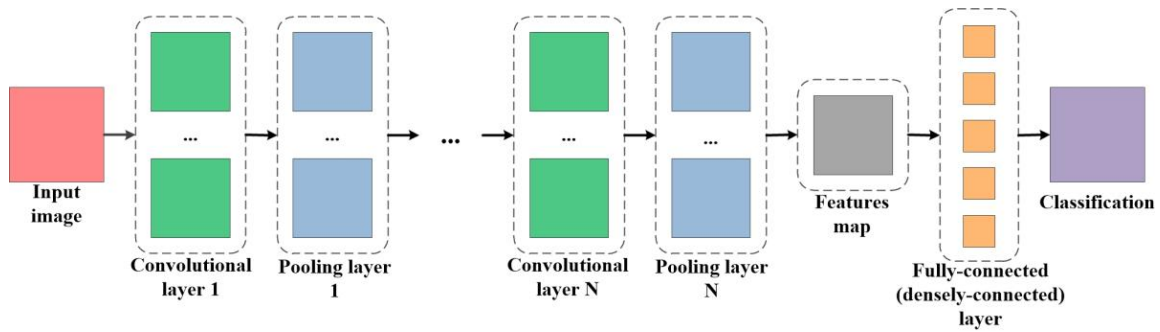

Figure 2: General architecture of a deep convolutional neural network.

## Implementation details

The default size of input image varies according to different DCNNs. Thus, two data augmentation approaches: image rotation and image rescaling are used to preprocess the image. First, all maps in *deepMap* are rescaled to the size fitting for a DCNN. Then, we randomly divided recalled maps into training dataset and test dataset. For the training dataset, image rotation creates three new maps by rotating the original one through 90, 180, and 270 degrees, respectively.

We have selected state-of-the-art DCNN models to test their performance of map-type classification. Each DCNN model is listed below.

- *AlexNet* (Krizhevsky, Sutskever and Hinton, 2012)

- *VGG Net* (Simonyan and Zisserman, 2014)
- *GoogleNet or Inception* (Szegedy, et al., 2015)
- *ResNet* (He, et al., 2016)
- *Inception-ResNet* (Szegedy, et al., 2017).

## Results and Discussions

Each DCNN has some influential improvements and designs. The objective of our experiments is to test whether and how much these improvements and designs effectively facilitate automatic map-type classification. Since DCNNs are sensitive to the quality and amount of training data, we make classifications according to different ratios of training data and test data. Table 1 lists the details of the experimental design and experimental results.

Table 1: Experimental design and results for comparing various (D)CNN methods.

| *CNNs & DCNNs* | *Experimental results* | |
|---|---|---|
| | *Group 1:* 60% data used for training & 40% data used for testing | *Group 2:* 80% data used for training & 20% data used for testing |
| AlexNet | *71% ~ 78%* | *77% ~ 83%* |
| VGG Net-19 | *73% ~ 80%* | *78% ~ 84%* |
| Inception V4 | *82% ~ 87%* | *88% ~ 94%* |
| ResNet V2-152 | *82% ~ 86%* | *89% ~ 93%* |
| ResNet-Inception V2 | *88% ~ 92%* | *95% ~ 99%* |

The experimental results have shown that the classification accuracies generated by these CNNs and DCNNs ranged from around 70% to 98%. Moreover, increasing the volume of training data would significantly raise the performance of DCNNs in map-type classification. This phenomenon supports the claim that it is critical to prepare large-scale well-labeled data to feed a neural network for enhancing its capability to distinguish different classes (Bengio, Courville and Vincent, 2012).

In detail, although AlexNet has been replaced by many later DCNNs in image classification, this pioneering CNN still enables the production of an acceptable result in map-type classification. The VGG method resulted in higher accuracy than did AlexNet, which supports the claim that the depth of a neural network is much more crucial than its spatial dimensions (Szegedy, et al., 2015). Moreover, GoogleNet or Inception organizes a very deep architecture of DCNN, which markedly improves classification accuracy and computational efficiency. However, the very deep network may produce higher accuracy results, but training is very difficult for a DCNN with a very deep architecture. ResNet proposed residual blocks to revolutionize the trade-off between efficient training and deep architecture in DCNNs. The results produced by ResNet prove that this strategy is also useful to facilitate map-type classification. To maintain the advantages of deep neural network and computational efficiency, Inception-ResNet integrates two compelling networks, inception network (Inception) and deep residual network (ResNet), as a unified and simplified architecture. This integrated DCNN produced the highest accuracy in image classification when a large-scale training dataset is available. Generally speaking, DCNNs can support the automated classification of the majority of map types in *deepMap*. However, the classification accuracy varies according to different map types. Some types of maps, such as the National Map's topographic maps, are difficult to distinguish without a well-labeled dataset.

Besides deep learning techniques, knowledge of map design and map generation are potential means to facilitate automatic map-type classification. Recently, the significance of transferring knowledge has been shown to substantially improve the performance of DCNNs by some edge-cutting models like NASNet (Zoph et al. 2017) and PNASNet (Liu et al., 2017). In the future, a DCNN that supports the transfer of knowledge, and high-level features of different maps will be our focus. We hope our work can contribute to the implementation of deep learning techniques in the cartographical community, and advance the progress of GeoAI.

**Xiran Zhou**, PhD student, School of Geographical Sciences & Urban Planning, Arizona State University, Tempe, AZ 85281

**Wenwen Li**, Associate Professor, School of Geographical Sciences & Urban Planning, Arizona State University, Tempe, AZ 85281

**Samantha T. Arundel**, Research Geographer, Center of Excellence in Geographic Information Science, U.S. Geological Survey, Rolla, MO 65401

**Jun Liu**, Associate Researcher, Shenzhen Institutes of Advanced Technology, Chinese Academy of Sciences, Shenzhen, Guangdong Province, P.R. China 518055